\documentclass[letterpaper, 10 pt, conference]{ieeeconf}  %

\IEEEoverridecommandlockouts                              %

\usepackage{graphicx}
\usepackage{amsmath,amsfonts}
\usepackage{afterpage}
\usepackage{xcolor}
\usepackage{booktabs}
\usepackage{comment}
\usepackage{caption}
\usepackage{cite}
\newcommand{\algoname}{OmniShape}

\newcommand{\realNum}{\mathbb{R}}
\newcommand{\coord}{\mathbf{p}}

\newcommand{\objectpose}{\mathbf{x}}
\newcommand{\triplane}{\mathbf{z}}
\newcommand{\triplanesAll}{\mathcal{Z}}
\newcommand{\fieldValue}{\mathbf{f}}
\newcommand{\latentDim}{n}
\newcommand{\lod}{p}

\newcommand{\interpolatedLatent}{\bar{z}}
\newcommand{\sethree}{SE(3)}
\newcommand{\image}{\mathbf{I}}
\newcommand{\normals}{\mathbf{N}}
\newcommand{\imagedim}{d}
\newcommand{\norf}{\mathbf{m}}
\newcommand{\projectedNORF}{\mathbf{\overline{m}}}
\newcommand{\sdfPoint}{\mathbf{s}}
\newcommand{\sdfDistance}{d}
\newcommand{\TVweight}{\alpha_{TV}}

\newcommand{\numObjects}{O}
\newcommand{\numSDFPCD}{M}

\newcommand{\diffusionState}{\mathbf{u}}
\newcommand{\diffusionParams}{\theta}
\newcommand{\diffusionTS}{t}

\newcommand{\diffusionNoise}{\epsilon}

\newcommand{\diffusionFunction}{\epsilon_\theta}
\newcommand{\diffusionShape}{\epsilon^z_\theta}
\newcommand{\diffusionNORF}{\epsilon^m_\theta}

\newcommand{\decoder}{\xi}

\newcommand{\interpolation}{\omega}

\title{\LARGE \bf
OmniShape: Zero-Shot Multi-Hypothesis Shape and \\ Pose Estimation in the Real World}

\author{Katherine Liu$^{1}$,
Sergey Zakharov$^{1}$, 
Dian Chen$^{1}$, 
Takuya Ikeda$^{2}$, \\
Greg Shakhnarovich$^{3}$,
Adrien Gaidon$^{1}$,
Rares Ambrus$^{1}$%
\thanks{$^{1}$Toyota Research Institute, Los Altos, CA 94022, USA. \{firstname.lastname\}@tri.global, $^\dagger$\{firstname.lastname\}.ctr@tri.global}
\thanks{$^{2}$Woven by Toyota, Chuo City, Tokyo 103-0022, Japan. 
\{firstname.lastname\}@woven.toyota}%
\thanks{$^{3}$Toyota Technological Institute at Chicago, Chicago, IL 60637, USA. 
\{firstname\}@ttic.edu \newline © 2025 IEEE.  Personal use of this material is permitted.  Permission from IEEE must be obtained for all other uses, in any current or future media, including reprinting/republishing this material for advertising or promotional purposes, creating new collective works, for resale or redistribution to servers or lists, or reuse of any copyrighted component of this work in other works.}%
}

\begin{document}

\maketitle
\thispagestyle{empty}
\pagestyle{empty}

\begin{abstract}
We would like to estimate the pose and full shape of an object from a single observation, without assuming known 3D model or category. In this work, we propose \algoname{}, the first method of its kind to enable probabilistic pose and shape estimation. \algoname{} is based on the key insight that shape completion can be decoupled into two multi-modal distributions: one capturing how measurements project into a normalized object reference frame defined by the dataset and the other modelling a prior over object geometries represented as triplanar neural fields. By training separate conditional diffusion models for these two distributions, we enable sampling multiple hypotheses from the joint pose and shape distribution. OmniShape demonstrates compelling performance on challenging real world datasets. Project website: \texttt{https://tri-ml.github.io/omnishape}
\end{abstract}

\section{Introduction}
\begin{figure}[!t]
  \centering
  \includegraphics[width=\columnwidth, trim={0.5cm 19cm 8cm 1cm}, clip]{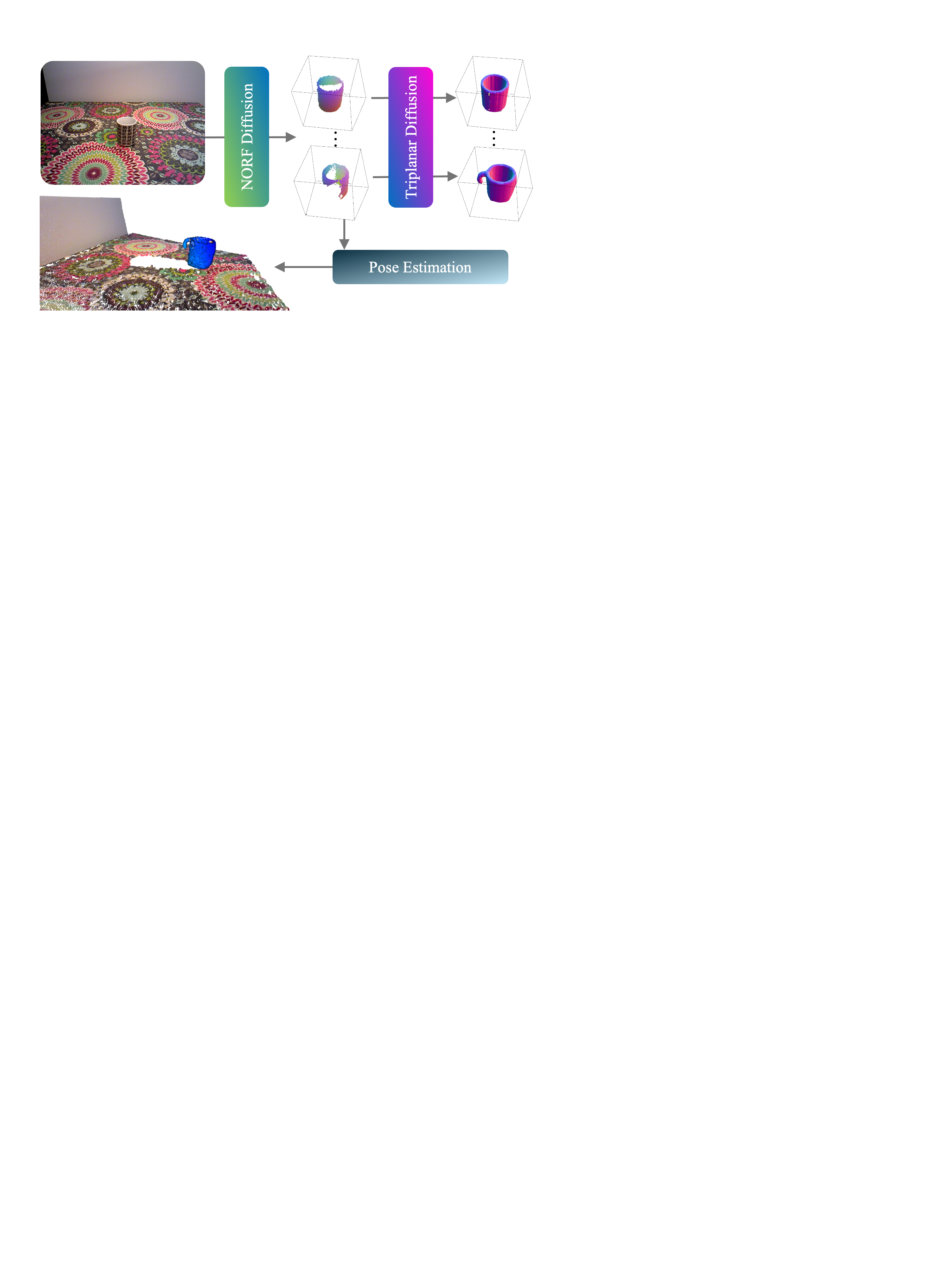}
  \caption{
\textbf{OmniShape} decomposes object estimation from an RGB-D image into two stages: estimating partial pointclouds in a normalized object reference frame and conditional shape completion. By using diffusion models for each stage, OmniShape predicts multiple hypotheses of pose and shape, in this case capturing possible object symmetry and/or canonicalization as well as potential geometry under occlusion. The first stage provides per-point association to depth images, enabling registration of the object. Input image from \cite{hodan2018bop}.}
  \label{fig:teaser}
\end{figure}

Detailed understanding of the 3D world is a core challenge in applications ranging from augmented reality to robotics. Despite recent progress in open-world image understanding \cite{gu2021open, kirillov2023segment}, estimating the complete and accurate 3D geometry of objects in a scene from a single view is an open problem. Consider the case of the cup in Fig. \ref{fig:teaser} for which a handle has not been observed: in such a case both the shape and pose of the object are uncertain and under-constrained. We are interested in enabling joint multi-hypothesis pose estimation \emph{and} shape completion. Our work is to our knowledge the first to address these two goals jointly, without assuming known geometry or tight constraints on object category.

The vast majority of techniques for \textbf{pose estimation} assume object geometry is known \textit{a priori} at an instance or category level. Given an observation of a scene comprised of known object models, the relative poses of the objects can be estimated with approaches such as classical correspondence based methods \cite{umeyama1991least, gao2003complete}, template matching \cite{nguyen2023gigapose}, inverse rendering \cite{yen2021inerf} and learning-based methods \cite{zakharov2019dpod}. A number of approaches use probabilistic techniques \cite{ikeda2024diffusionnocs,deng2021poserbpf, zhang2024generative} to deal with pose uncertainty stemming from self-occlusion and symmetry. However, assumptions of known object or limited categories severely limits existing methods' utility in open-world settings.
Some recent methods \cite{park2020latentfusion, wen2023bundlesdf} estimate the pose of novel instances, but assume multiple observations.

While single-view \textbf{shape completion} methods seek to estimate full 3D object geometry, existing approaches make assumptions that make them difficult or brittle to apply in real-world estimation tasks. For example, many methods use the ShapeNet dataset\cite{chang2015shapenet}, where instances within a single class are aligned, to learn shape representations \cite{park2019deepsdf, cheng2022sdfusion}. Other methods eschew known canonicalization, learning image-conditioned generative models \cite{radford2021learning, nichol2022point} over large-scale data. However, both such approaches can be challenging to apply in the real world as their internal reference frames are difficult to relate to metric observations. An alternative is to assume objects are observed in identity pose and complete object shapes in the camera coordinate frame, featuring methods including regression \cite{huang2023zeroshape, wu2023multiview} and novel-view synthesis \cite{liu2023one, tang2023dreamgaussian}. Such approaches define bounds on the extents of the objects for surface extraction, which can be brittle depending on the severity of self-occlusion.

\begin{figure*}[!h]
\centering
\vspace{1mm}
    \includegraphics[width=0.95\textwidth, trim={0.1cm 20.8cm 0.1cm 0cm}, clip]{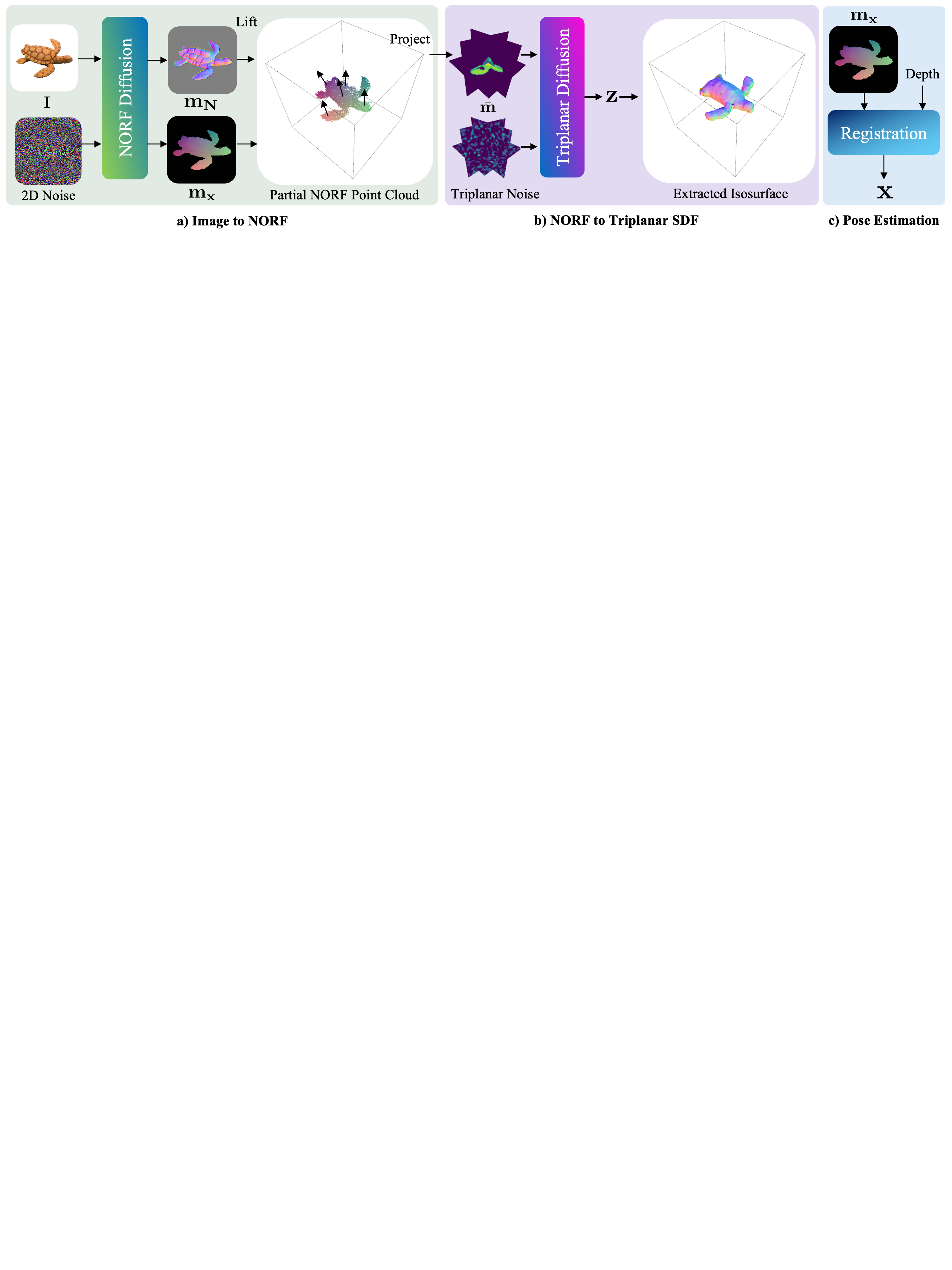}
    \caption{\textbf{System Overview.} \algoname{} decouples shape completion into two generative stages. The first (a) maps an RGB (with optional normals from a depth image) to a partial pointcloud with corresponding normals in a NORF. The second (b) conditions on the partial observation in the NORF and predicts the complete object geometry (shown here with CFG), represented as a triplane. Both stages are modeled with diffusion models to produce multiple hypotheses. (c) Given additional information such as a depth image, the object can be registered into the scene using the first stage output. A reduced number of illustrative normals in the NORF pointcloud and channels in the Ortho-NORF conditioning drawn for visual clarity.
    }
    \label{fig:system_diagram}
\end{figure*}

In this work, we propose \algoname{}, a novel framework for joint shape completion and pose estimation in the real world. \algoname{} is based on the key technical insight that shape completion can be decoupled into two multi-modal distributions.
Relaxing the strict dataset canonicalization assumptions of previous work \cite{wang2019normalized, ikeda2024diffusionnocs}, the first model captures the mapping between images and a Normalized Object Reference Frame (NORF), implicitly capturing pose and partial shape. The second model learns a conditional distribution over complete object shapes, represented as triplanar neural fields.
By learning a distribution over NORFs, OmniShape generates partial estimates to condition shape completion in a normalized reference frame, rather than requiring partial measurements to be arbitrarily normalized into a fixed coordinate system. 
We model the rich multi-modal nature of the distributions via Denoising Diffusion Probabilistic Models \cite{ho2020denoising}. \algoname{} outputs pairs of shapes and dense correspondences that enable placing the predicted object into the scene, bridging probabilistic pose estimation and generative shape modeling.
We train \algoname{} on synthetic data and test on challenging real-world estimation tasks.

\textbf{In summary, our contributions are as follows:}
(1) A multi-hypothesis algorithm for jointly estimating the pose and complete shape of objects from a single image, without requiring any prior knowledge about the object
(2) A framework that applies the notion of predicting normalized object coordinate spaces \cite{wang2019normalized} from images to ``in-the-wild" datasets, relaxing strict canonicalization requirements for use in single-view shape completion
(3) An approach to shape completion via diffusion of objects modelled as triplanar grids conditioned on a partial pointcloud observation.

\section{Our Approach:~\algoname{}}

We would like to jointly estimate the pose $\objectpose \in \sethree$ and shape $\triplane$ (described in Sec. \ref{sec:triplanar_fields}) of an object from a single cropped, segmented RGB-D observation $\image \in \realNum^{\imagedim \times \imagedim \times 4}$, where $\imagedim$ is the crop resolution. Generally, an object's pose cannot be fully specified without knowledge of the shape, and vice versa. However, the joint conditional probability distribution $p(\objectpose,\triplane|\image)$ can be difficult to sample directly from due to its complex and multi-modal nature. 

Inspired by prior work in canonical coordinate estimation~\cite{wang2019normalized}, we replace $\objectpose$ with an image-like map $\norf \in \realNum^{\imagedim \times \imagedim \times 3}$ that projects normalized 3D coordinates of visible object points to the camera reference frame -- the NORF map (Sec.~\ref{sec:norf}). The NORF map provides dense pixel to 3D association, enabling the recovery of $\objectpose$ from $\norf$ via registration methods given observed depth.
We then model the joint probability over object geometry and pose as $p(\triplane, \norf | \image)$. 

Besides enabling pose estimation, $\norf$ \emph{also} provides a partial pointcloud observation of the object surface, from which the object shape can be completed. This is the key insight in~\algoname{}: we disentangle joint reasoning about pose and shape into a chain of two distributions: 
(1) the observed surface points in a normalized object reference frame $\norf$  given the image $\image$ and (2) the object geometry $\triplane$ given the partial observation in $\norf$:
\begin{equation}\label{eq:product}
    p(\triplane,\norf|\image) = p(\triplane|\norf)p(\norf|\image),
\end{equation}
where we assume that $\norf$ provides the necessary information to model $\triplane$.
\algoname{} approximates both conditional distributions with diffusion models, learning two models $\diffusionNORF$ and $\diffusionShape$ to represent $p(\norf|\image)$ and $p(\triplane|\norf)$ respectively, and enabling sampling from the joint distribution via~\eqref{eq:product}. We highlight that a depth measurement is only needed at test time to estimate a scaled, metric pose.

\subsection{Diffusion Preliminaries}\label{sec:diffusion}
Denoising Diffusion Probabilistic Models (DDPMs) \cite{ho2020denoising} model generative processes by learning to iteratively denoise noisy inputs. 
Intuitively, diffusion models assume a forward noising process of iteratively adding normally distributed noise to the state $\diffusionState$: $q(\diffusionState_\diffusionTS | \diffusionState_{\diffusionTS-1}) = \mathcal{N}(\sqrt{1-\beta_\diffusionTS}\diffusionState_{\diffusionTS-1}, \beta_\diffusionTS \mathbf{I})$, where $\beta_\diffusionTS$ changes according to a predefined variance schedule.
To enable a backwards ``denoising" process, a function $\diffusionFunction$ can be trained to predict the amount of unscaled noise $\diffusionNoise \sim \mathcal{N}(0,\mathbf{I})$ in a given noisy input $\diffusionState_\diffusionTS$, i.e., to minimize a noise matching objective \cite{ho2020denoising}:
\begin{equation}
\mathcal{L}(\diffusionParams) = \mathbb{E}_{t, \diffusionState_0, \epsilon}\left[\left\|\epsilon - \diffusionFunction(\diffusionState_\diffusionTS, \diffusionTS)\right\|^2\right].
\label{eq:diffusion_objective}
\end{equation}
Given a trained denoising function, a sample drawn from random noise is then iteratively denoised.
Diffusion models can be further made to model conditional distributions via methods such as classifier guidance \cite{dhariwal2021diffusion} or classifier-free guidance (CFG)\cite{ho2022classifier}. 
In this work, we approximate both $p(\triplane|\norf)$ and $p(\norf|\image)$ with diffusion models, and optionally use classifier free guidance to generate samples from $p(\triplane|\norf)$. 

\subsection{NORF Diffusion}\label{sec:norf}
The Normalized Object Coordinate Space (NOCS)~\cite{wang2019normalized} maps pixel coordinates to a normalized reference frame for pose estimation. This requires \emph{canonicalization} -- alignment to a shared coordinate system, e.g., defined for a coherent semantic category. However, we are interested in more general scenarios and datasets, where  canonicalization cannot be ensured, such as the recently proposed Objaverse dataset \cite{deitke2023objaverse}. 
We therefore relax this assumption, and to avoid confusion, name our normalized coordinate framework NORF: Normalized Object Reference Frame. Other methods have also recognized the utility of NOCS-like parameterizations for shape estimation from multi-view images \cite{xu2024sparp} and text \cite{li2023sweetdreamer}; \algoname{} utilizes this insight specifically for decoupling multi-hypothesis shape and pose estimation.

\algoname{} assumes a dataset of $\numObjects$ object models, each contained in a unit cube centered at the origin of the 3D coordinate system. Following \cite{wang2019normalized}, we project the visible surface into a posed camera with known intrinsics to obtain a NORF map $\norf_x \in \realNum^{\imagedim \times \imagedim \times 3}$, an image-like quantity where each pixel value indicates the 3D position in the NORF. We also build a NORF normal map $\norf_N \in \realNum^{\imagedim \times \imagedim \times 3}$, where each pixel value is the surface normal of the observed point. We construct training tuples of observed images, observed normals, and corresponding output partial NORF maps, i.e., $(\image, \normals, \norf)$, where $\norf_x$ and $\norf_N$ comprise the NORF measurement $\norf$.
As in previous work \cite{ikeda2024diffusionnocs} we use normal $\normals$, which can be calculated from depth, to avoid brittleness in normalizing arbitrary depth measurements. 
We train the DDPM as in~\eqref{eq:diffusion_objective}, with the NORF map $\norf$ as state, and the observed RGB image $\image$ and normals $\normals $ as the conditoning:
\begin{equation}\label{eq:diffusion_norf}
\mathcal{L}(\theta) = \mathbb{E}_{t, (\norf_0,\image,\normals), \epsilon}\left[\left\|\epsilon - \diffusionNORF(\norf_t, t, \image, \normals)\right\|^2\right].
\end{equation}
We use $\diffusionNORF$ to approximate conditional sampling of partial pointcloud observations $p(\norf|\image)$, illustrated in Fig. \ref{fig:system_diagram}a.

Although \algoname{} does not enforce strict intra-class canonicalization, our experiments show the sampled NORF maps trained on human generated or processed datasets exhibit evidence of structure. Fig. \ref{fig:ocrtoc} shows  cup openings pointing upwards and the axis aligned airplanes, suggesting some inherent alignment rules. The canonicalization and therefore pose distribution modeled is a function of the dataset. \algoname{} uses these patterns to avoid optimizing triplanes online for objects in arbitrary poses as shapes can be completed in common reference frames learned from data.

\subsection{Triplanar Field Diffusion} \label{sec:triplanar_fields}
We model objects as signed distance fields, represented by triplanar neural fields \cite{chan2022efficient}. 
Each object is represented by a triplanar latent $\triplane \in \realNum^{3 \times 2^\lod \times 2^\lod \times \latentDim}$, where $\latentDim$ is the dimension of the latent and $\lod$ is the level of detail. The triplanar representation allows for continuous neural fields to be represented via three orthogonal $2^\lod \times 2^\lod$ feature planes. To query for the signed distance of an arbitrary point $\coord \in \realNum^3$, we project the coordinate onto three orthogonal planes. 
We then perform interpolation per plane and concatenate the resulting features to obtain the latent for the coordinate, i.e., $\interpolatedLatent_{\coord} = \interpolation (\coord, \triplane)$, where $\interpolatedLatent \in \realNum^{3n}$. To obtain the final signed distance value $\fieldValue_\coord$, we learn a decoder $\decoder$ such that $\fieldValue_\coord = \decoder(\interpolatedLatent_\coord)$. 

For each object, we assume a SDF pointcloud tuple $\{ (\sdfPoint^i_0, \sdfDistance^i_0) ... (\sdfPoint^i_{\numSDFPCD_i}, \sdfDistance^i_{\numSDFPCD_i}) \}$, which pairs ${\numSDFPCD_i}$ sampled 3D points $\sdfPoint \in \realNum^3$ coupled with their distance $\sdfDistance \in \realNum$ from the surface of the $i$-th object. 
We formulate an optimization over the set of triplanar latents $\triplanesAll = \{\triplane_0, ..., \triplane_\numObjects\}$ as well as the set of parameters of the decoder $\decoder$ to minimize the L1 reconstruction loss, combined with a total variation (TV) term summed over each of the three feature planes as in~\cite{shue20223d}:
\begin{equation}
\mathcal{L}(\triplanesAll,\decoder) = \sum_{i=0}^{i=\numObjects} \sum_{j=0}^{j=\numSDFPCD_i} \left| \decoder(\interpolation (\sdfPoint^i_j, \triplane_i)) - \sdfDistance_j^i \right| + \TVweight \sum_{i=0}^{i=\numObjects} \mathrm{TV}(\triplane_i).
\end{equation}
After a set of triplanes has been optimized, we use pairs of optimized triplanes and partial pointclouds with normals in the NORF to train the shape completion diffusion model. The triplanar representation can be rearranged into image-like tensors of dimension $\triplane'\in\realNum^{2^\lod \times 2^\lod \times 3\latentDim}$, allowing the use of 2D diffusion methods for the shape completion model $\diffusionShape$. Importantly, the shape completion process can assume that NORF predictions are \textit{already} in a normalized reference frame, avoiding brittle pre-prediction normalization.

We find it is important at the shape completion stage to align the NORF predictions described in Sec. \ref{sec:norf} to the trained triplanes. To this end, we voxelize $\norf$ into an occupancy grid with side dimension $2^{\lod + 1}$ (keeping the average normal value for occupied cells), then orthogonally project the values onto the three planes, generating measurements aligned with the triplanar representation. We then perform Pixel Unshuffling \cite{shi2016real} to reduce the spatial dimension of the conditioning by half.
This process builds the Ortho-NORF $\projectedNORF\in\mathbb{R}^{{2^\lod }\times{2^\lod }\times48}$: the partial pointcloud with normals, orthogonally projected into the triplanar space.
At inference time, we filter noisy points from predicted $\norf$ before projection.

We then adapt Eq.~\ref{eq:diffusion_objective} for shape completion diffusion:
\begin{equation}
\mathcal{L}(\theta) = \mathbb{E}_{t, (\triplane', \projectedNORF), \epsilon_t}\left[\left\|\epsilon_t - \diffusionShape(\triplane'_t, t, \projectedNORF)\right\|^2\right],
\end{equation}
where the state is a triplane $\triplane'$ from $\triplanesAll$ and the conditioning is $\projectedNORF$. At inference time, we sample from $p(\triplane|\norf)$ using the trained $\diffusionShape$.
Fig. \ref{fig:system_diagram}b illustrates this process.

\subsection{Shape Completion and Registration}
Given an RGB-D image with an object segmentation mask at inference time, %
\algoname{} uses the diffusion model $\diffusionNORF$ to sample $\norf$, which is used as conditioning to sample object completions from the trained model $\diffusionShape$. From the sampled $\norf$ and $\triplane$, an explicit pose $\objectpose$ and surface can be extracted, enabling the object to be placed into the scene. Multiple hypotheses from the joint distribution of pose and shape can be generated by applying the iterative sampling process multiple times.
Registration is enabled by the per-pixel association between a point in the NORF and the observed image given by $\norf$.
We show in Sec. \ref{sec:experiments} that a pose registration metric can also be a practical hypothesis selection method.
\begin{figure*}[!h]
\centering
\vspace{1mm}
      \includegraphics[width=0.95\textwidth, trim={0.1cm 20cm 0.1cm 0cm}, clip]{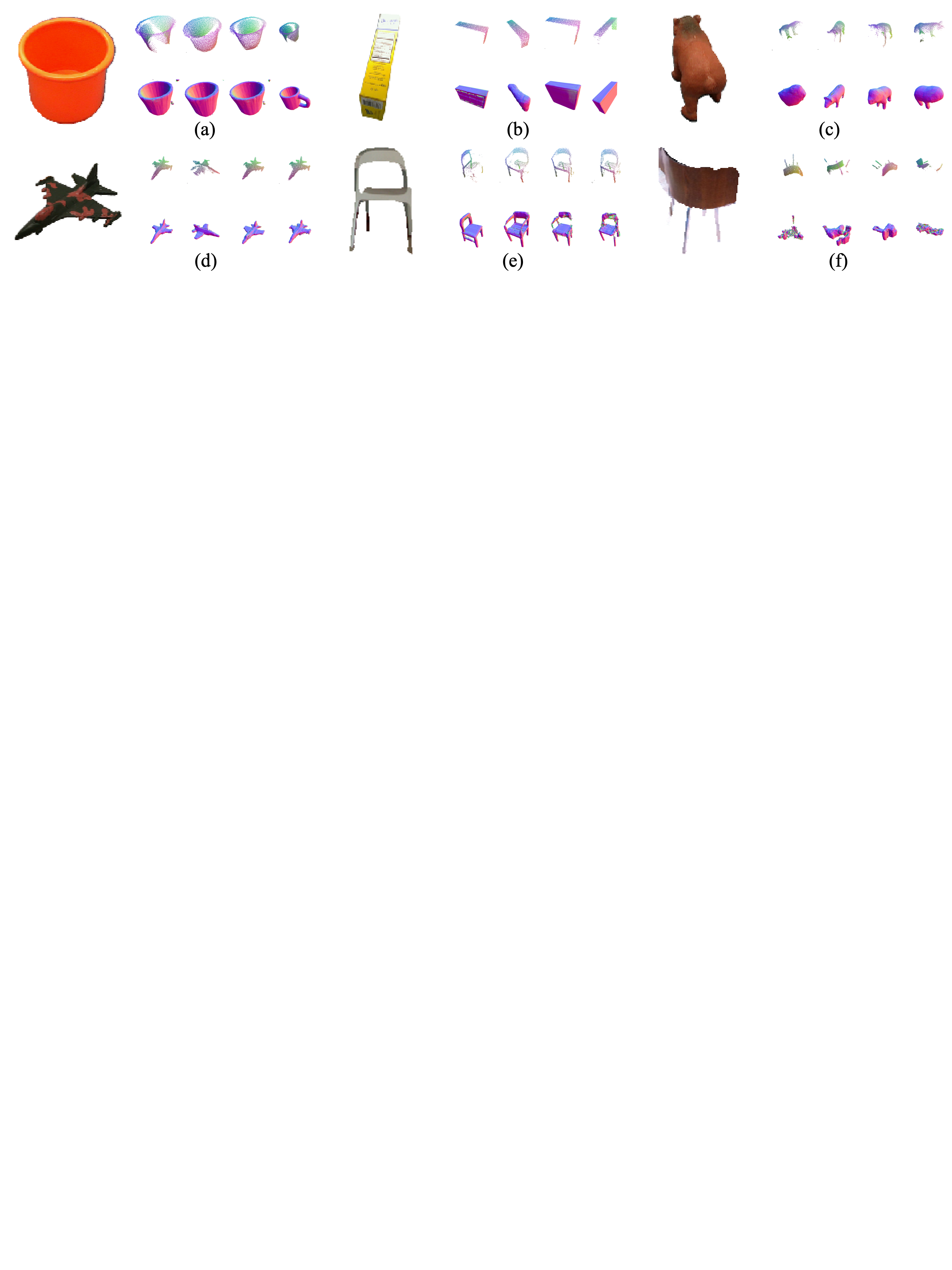}
    \caption{\textbf{Qualitative shape completion results.} Given an RGB image, \algoname{} estimates multiple hypotheses for both the partial pointcloud in the NORF and the corresponding shape completion. For each input image, we show four partially filtered partial pointcloud predictions (top row) and one conditional shape completion each (bottom row). The probabilistic nature of \algoname{} enables the prediction of different potential shape completions, useful under ambiguity. For example, \algoname{} predicts cups of different geometries in (a), boxes of different dimensions in (b), and different animals in (c).  In (f) we show an example failure case exhibiting incoherent geometry. (a)-(d) from Ocrtoc3D and (e)-(f) from Pix3D. }
    \label{fig:ocrtoc}
\end{figure*}

\section{Experiments}
\label{sec:experiments}
To evaluate \algoname{}, we test performance on several challenging object-centric estimation datasets.

\subsection{Implementation Details}
\textbf{Data.}
We train \algoname{} on the dataset proposed in ZeroShape \cite{huang2023zeroshape}, featuring 84789 objects with corresponding SDF supervision and over 1 million corresponding renderings sourced from ShapeNet \cite{chang2015shapenet} and the Objaverse-LVIS split \cite{deitke2023objaverse}. While $\diffusionNORF$ is trained with optional input normals, for all evaluation results we use only RGB images.

\textbf{Triplane Model.}
For our experiments, we use triplanes with $\lod = 5$ and $\latentDim = 12$, use a MLP with layers of width $(36, 512, 512, 1)$ and Leaky ReLU activations to decode interpolated latent values into signed distances. We set $\TVweight = 0.01$ and randomly sample 5k supervision points per object each epoch. The network is trained on eight NVIDIA A100s with a batch size of 128 per GPU and learning rate of $0.01$ for 9299 epochs. We use octree subdivision (similar to \cite{irshad2022shapo}) to a level of detail 6 to extract points on the isosurface. Before input to $\diffusionShape$ we normalize the triplanes per channel to have a standard deviation of $0.2$, and clip values to $[-1, 1]$.

\setlength{\tabcolsep}{4pt}
\begin{table}[!h]
\caption{\textbf{Single-object reconstruction results.}  \textit{Ours, FH} indicates results when taking the first hypothesis generated. \textit{Ours, Best-of-$N$} indicates taking the best result compared to the ground-truth according to the Chamfer distance over $N$ hypotheses, which we report to quantify the quality of the best hypothesis generated. We report our metrics as (without CFG/with CFG=5). 
\algoname{}'s first hypothesis is competitive with the baseline methods, while
taking additional hypotheses results in better solutions existing in the hypothesis set
indicating the benefits of our probabilistic formulation. All baseline metrics taken from ZeroShape \cite{huang2023zeroshape}.}  
\resizebox{\columnwidth}{!}{%
\begin{tabular}{lllll}
\toprule
 & \multicolumn{2}{c}{Ocrtoc3D}& \multicolumn{2}{c}{Pix3D}\\
 \cmidrule(lr){2-3} \cmidrule(lr){4-5}
 Method& FS@1↑  & CD↓     & FS@1↑  &CD↓     \\
\hline
SS3D \cite{alwala2022pre}       & 0.1271 & 0.543  & 0.1326 &0.485 \\
MCC \cite{wu2023multiview}        & \underline{0.1994}& 0.411  & 0.1754 &0.514 \\
One-2-3-45 \cite{liu2024one} & 0.1323 & 0.492  & 0.1364 &0.443 \\
OpenLRM \cite{openlrm}    & 0.1552 & 0.432  & 0.1458 &0.492 \\
Shap-E \cite{jun2023shap}     & 0.1725 & 0.395  & \textbf{0.2016} &\textbf{0.340} \\
ZeroShape \cite{huang2023zeroshape}  & \textbf{0.2410} & \textbf{0.286}  & \underline{0.1928}&\underline{0.345}\\
    Ours, FH& 0.1952/0.1975& 0.376/\underline{0.367}& 0.1675/0.1704&0.426/0.426\\
\hline
Ours, best-of-5& 0.2477/0.2491& 0.272/0.268& 0.2214/0.2200&0.318/0.326\\
Ours, best-of-25  & 0.2856/\textbf{0.2867}& 0.233/\textbf{0.230}& \textbf{0.2622}/0.2571&\textbf{0.263}/0.272\\
\bottomrule
\end{tabular}
}
\label{tab:ocrtoc}
\end{table}

\textbf{Diffusion Models.}
For $\diffusionNORF$, we use square crops of dimension 128 and a 2D UNet \cite{Ronneberger2015UNetCN} implementation \cite{von-platen-etal-2022-diffusers} with channel dimensions of $(128, 128, 256, 256, 512, 512)$, with the fifth layer of the encoder and second layer of the decoder as attention blocks. 
For $\diffusionShape$, we use a modified version of different UNet variant \cite{dhariwal2021diffusion}, with channel dimensions $(540, 1080, 2160)$ and attention at resolutions 2,4. Both networks assume a linear noise schedule with 1000 steps and beta ranging from 0.0001 to 0.02, and train using cosine learning rate decay with 500 warm up steps and peak learning rate of 0.0001 on eight NVIDIA A100s. $\diffusionNORF$ is trained with a batch size of 128 per GPU for 1000 epochs, while $\diffusionShape$ is trained with a batch size of 32 per GPU for 100 epochs.

\setlength{\tabcolsep}{3pt}
\begin{table*}%
\centering
\vspace{1mm}
\caption{\textbf{Real-world results}. On the challenging task of estimating the shape of objects and registering the objects in real-world coordinates, \algoname{} predictions paired with inlier selection over 10 hypotheses outperforms Zeroshape on TYO-L and NOCS as measured by L1 Chamfer Distance and F1 score (threshold of 5).
Taking multiple \algoname{} hypotheses generates estimates with higher accuracy across all three datasets. Due to high variance when taking the average over all objects, we instead report the mean and standard deviation over the per-scene means, bolding results based on mean value.}
\begin{tabular}{lccccll}
\toprule
                           & \multicolumn{2}{c}{\textbf{TYO-L} (21 scenes, 1670 objects in total)} & \multicolumn{2}{c}{\textbf{NOCS} (6 scenes, 875 objects in total)} &\multicolumn{2}{c}{\textbf{HOPE} (10 scenes, 920 objects in total)}\\
                           \cmidrule(lr){2-3} \cmidrule(lr){4-5} \cmidrule(lr){6-7}
                           Method
                           & CD↓         & F1↑       & CD↓            & F1↑            &CD↓             &F1↑            \\
\midrule
Zeroshape                  & 13.35$\pm$5.36& 0.31$\pm$0.10& 14.57$\pm$2.82& 0.30$\pm$0.02&\textbf{9.94}$\pm$4.48&0.49$\pm$0.06\\
Ours-RGB, first hypothesis& 10.98$\pm$4.89& 0.47$\pm$0.11& 15.45$\pm$2.26& 0.35$\pm$0.05&13.35$\pm$10.76&0.47$\pm$0.06\\
 Ours-RGB, inlier selection& \textbf{8.81}$\pm$4.04& \textbf{0.54}$\pm$0.11& \textbf{10.67}$\pm$1.22& \textbf{0.42}$\pm$0.04& 10.36$\pm$7.64&\textbf{0.54}$\pm$0.07\\
\hline
Ours-RGB, best-of-10& 5.76$\pm$2.56& 0.65$\pm$0.13& 7.73$\pm$0.87& 0.51$\pm$0.05&7.04$\pm$3.90&0.64$\pm$0.08\\
\bottomrule
\end{tabular}
\label{tab:bop}
\end{table*}

\begin{figure*} [thpb]
\centering
      \includegraphics[width=0.98\textwidth, trim={0.1cm 18.6cm 0.1cm 0cm}, clip]{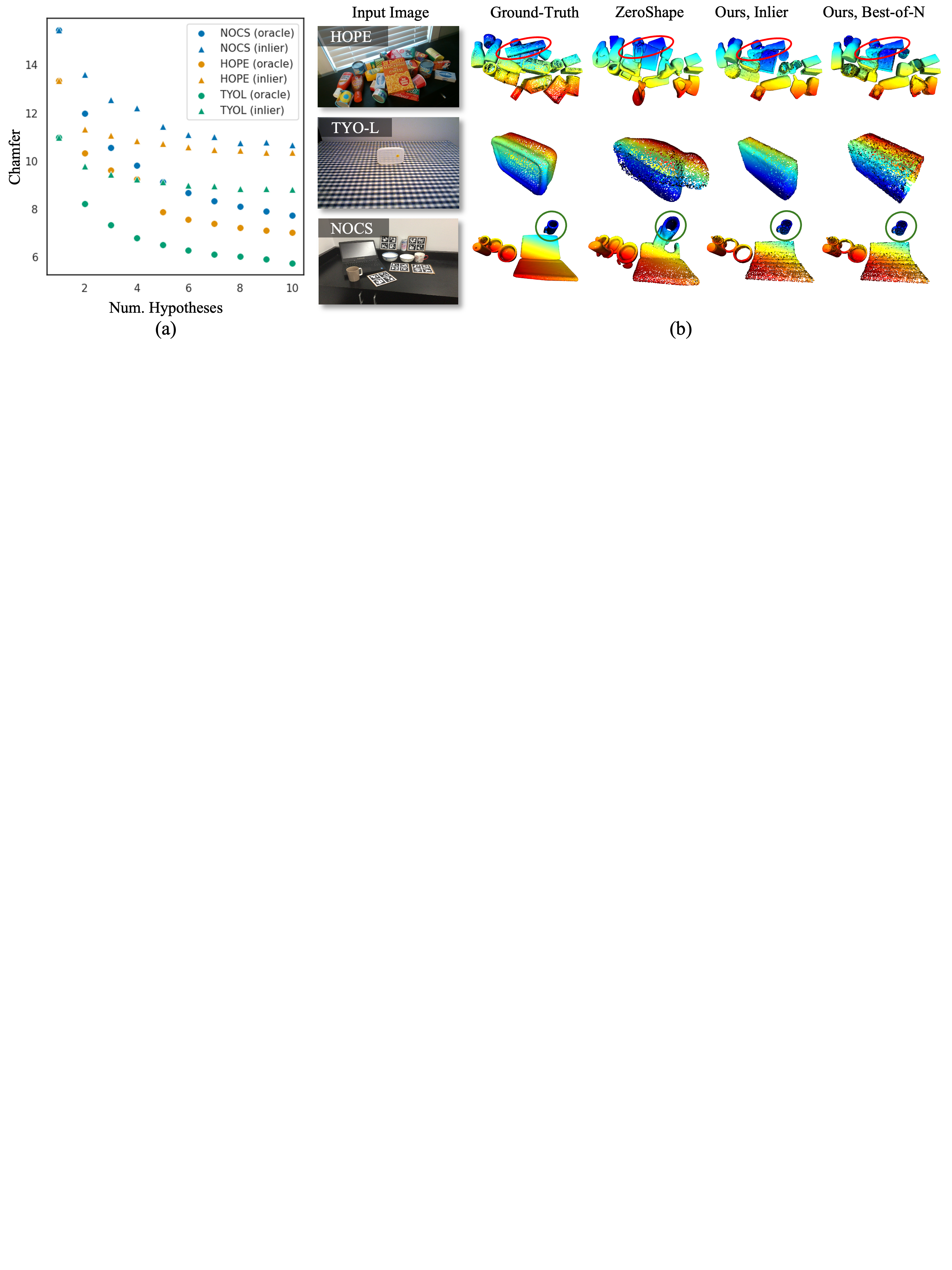}
    \caption{\textbf{Results on shape and pose estimation.} (a) We show that sampling additional hypotheses  
    reduces the Chamfer distance on all three datasets, with the inlier-based method plateauing earlier than the oracle-based method. (b) Qualitative visualizations of performance on the datasets. We show the input image as well as the groundtruth and reconstruction by various methods. The middle example (from TYO-L) highlights a key benefit of the multi-hypothesis nature of our method; OmniShape can predict shapes of varying sizes given an ambiguous view of only the front of the container. We also observe that although neither ZeroShape nor \algoname{} are trained to handle occlusions, \algoname{} can sometimes fill in missing geometry as in the cup on the bottom row (green circle) and the thin spaghetti box on the top row (red ellipse). Training time image augmentations may also help \algoname{} handle images encountered in the wild.}
    \label{fig:bop}
\end{figure*}

Conditioning is implemented via channel concatenation, except for timestep conditioning which depends on the implementations noted above. When training $\diffusionNORF$, we downscale-upscale with $25\%$ probability, rotate input and output with probability $50\%$, and drop normal conditioning with probability $50\%$; for $\diffusionShape$ we drop all conditioning with probability $20\%$. We do not use CFG except where noted in Tab. \ref{tab:ocrtoc} paired only with $\diffusionShape$, experimentally finding that CFG benefits performance on only some test sets.
For inference, we use the DPM-Solver++ \cite{lu2022dpm} with 50 steps for $\norf$ and 25 steps for $\triplane$. Without batching or CFG, generating a single Ortho-NORF estimate takes approximately $2.1$s seconds and a single shape completion (including surface extraction) takes approximately $0.9$s on a NVIDIA RTX A6000.

\textbf{Baselines.} We compare to several pre-trained zeroshot image-to-3D methods,
both deterministic: SS3D \cite{alwala2022pre}, MCC \cite{wu2023multiview}, One-2-3-45\cite{liu2024one}, LRM\cite{hong2023lrm}/OpenLRM\cite{openlrm}, ZeroShape \cite{huang2023zeroshape} and stochastic: Shap-E \cite{jun2023shap}. 
Object representations include NERFs \cite{alwala2022pre, liu2024one, hong2023lrm}, pointclouds \cite{wu2023multiview}, and occupancy fields \cite{huang2023zeroshape}.
SS3D learns a shape space using ShapeNet\cite{chang2015shapenet}, then utilizes image datasets via a multi-hypothesis camera approach.
One-2-3-45 leverages a novel view synthesis diffusion model \cite{liu2023zero} to inform shape completion. OpenLRM is an implementation of LRM \cite{hong2023lrm}, a transformer based method which predicts triplanar NeRFs trained on Objaverse. ZeroShape predicts depth and intrinsics, then normalizes the observed surface from which the object is completed. MCC is an encoder-decoder method trained on CO3D \cite{reizenstein21co3d}.
Shap-E is a diffusion-based method trained on millions of 3D objects \cite{mildenhall2021nerf} that predicts objects in an internal (unknown) reference frame.
ZeroShape and \algoname{} predict only geometry and are the only methods to share the same training split.

\begin{figure*}[!h]
\centering
\vspace{1mm}
      \includegraphics[width=0.78\textwidth, trim={0.4cm 21.3cm 0.4cm 0.1cm}, clip]{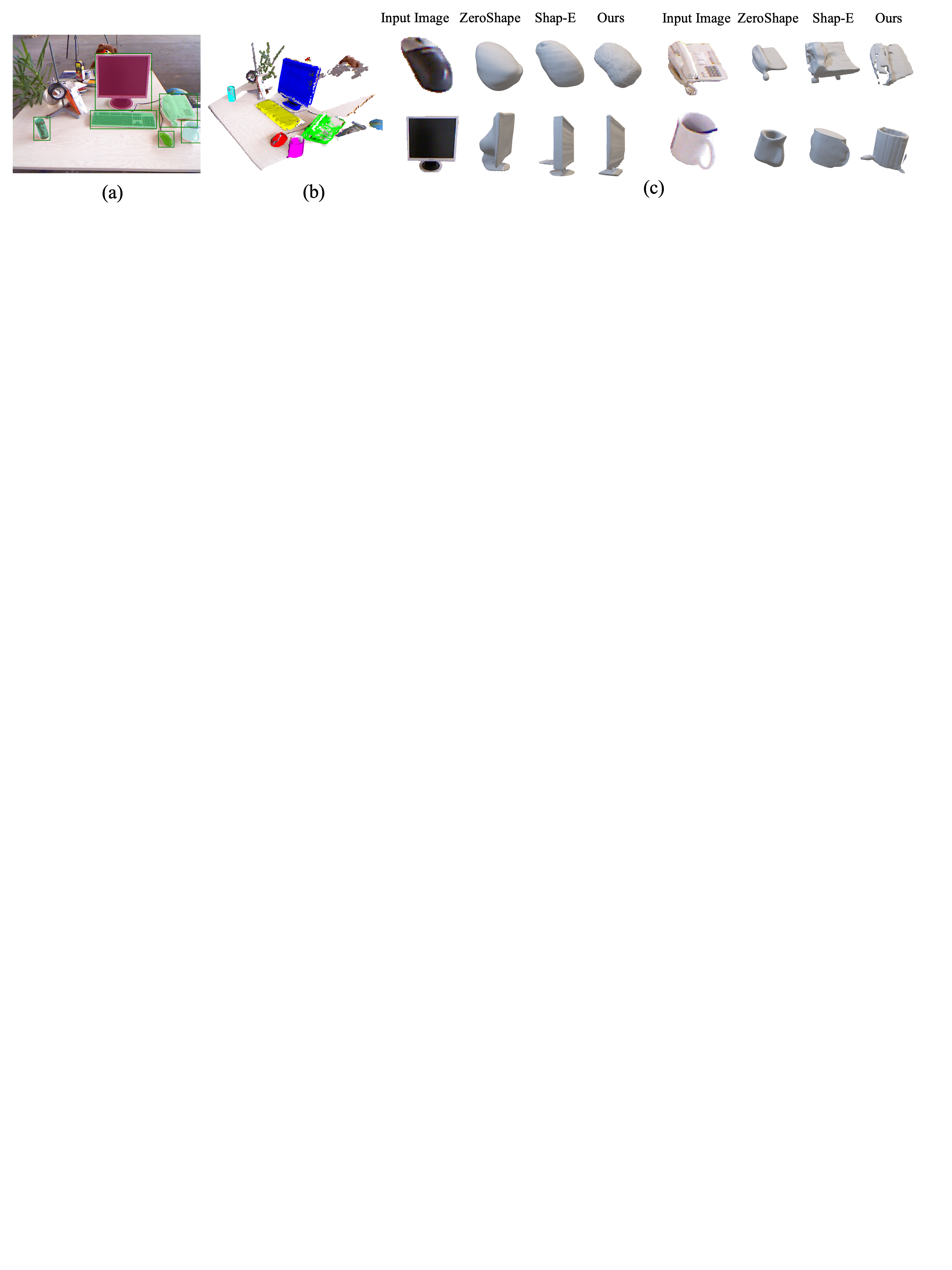}
    \caption{\textbf{Qualitative example of shape and pose estimation.} We use the inlier hypothesis selection method over 25 OmniShape hypotheses to complete objects in an image from the TUM dataset \cite{sturm12iros}. (a) We detect \cite{liu2023grounding} and segment \cite{ravi2024sam2} several objects from the scene and show (b) OmniShape meshes (extracted via a differentiable watertight mesh extractor \cite{wei2023neumanifold} and decimated) and estimated NORF points after being registered into the scene, overlaid on the input pointcloud. In (c), we show qualitative performance of \algoname{}, ZeroShape and Shap-E. Each method estimates objects in a different reference frame; we manually rotate to provide approximate representative views. Videos of meshes are provided in the attached multimedia.}
    \label{fig:inthewild}
\end{figure*}

\subsection{Single-Object Reconstruction}
We evaluate shape completion performance on two real world datasets of single objects, using the processed versions provided by ZeroShape \cite{huang2023zeroshape}:
Ocrtoc3D\cite{shrestha2022real}, consisting of 749 images of diverse objects captured in the wild, and Pix3D \cite{sun2018pix3d}, featuring 1181 images of primarily furniture. We report Chamfer distance and F1 score following the protocol described in \cite{huang2023zeroshape}, finding the lowest Chamfer distance over a discrete set of rotations, using 10k points from the estimate.

Quantitative results are given in Tab. \ref{tab:ocrtoc} and selected qualitative results in Fig.~\ref{fig:ocrtoc}. Baseline values (all except ours) are taken from \cite{huang2023zeroshape}.
Given the first hypothesis, \algoname{} exhibits performance similar to several state of the art methods.
Taking the best estimate of 5 \algoname{} hypotheses (as determined by comparing to the ground-truth) our method outperforms the other methods and further improves with 25 hypotheses.
Although the best-of-N metric cannot be used in online inference, these results demonstrate the usefulness of a multi-hypothesis method. Furthermore, unlike the other generative method considered \cite{jun2023shap} that predicts objects in an unknown coordinate system,
\algoname{} enables the predicted shape to be estimated in world coordinates.

\subsection{Real-World Reconstruction and Pose Estimation}

We evaluate \algoname{} for object shape and pose estimation on three real-world datasets, using the provided images, depths, segmentations, and meshes. 
Image inputs are interpolated via a nearest strategy to the size required by each method.
TYO-L \cite{hodan2018bop} features one object per scene with varied poses/lighting; we keep a max of the first 80 images from each of the 21 scenes. NOCS \cite{wang2019normalized} includes multiple objects such as mugs and laptops; we keep the first 25 images from each of the 6 scenes in the REAL275 test set. HOPE \cite{tyree2022hope} includes crowded toy food scenes; we keep the first 5 images from each of the 10 scenes.
These benchmarks are typically used to evaluate pose estimation with strictly consistent object canoncalizations between train and test. OmniShape relaxes strict canonicalization requirements but therefore cannot directly measure object pose error compared to an arbitrary canonical frame defined by any one benchmark. Instead, we measure performance via Chamfer-L1 and F1 score after object registration into the scene, sampling 10k points from estimated objects.

We compare to ZeroShape \cite{huang2023zeroshape}, one of the best performing baselines from Tab. \ref{tab:ocrtoc} that also provides a straight-forward method of estimating the object pose in metric coordinates. Procrustes and RANSAC are used to align the observed depth and prediction. For ZeroShape we solve for the alignment between the predicted pointcloud from its first stage to the points observed from depth, while for \algoname{} we use the estimated $\norf_x$, which provides the 3D NORF coordinate of a given pixel.
We use the same registration settings for both methods, including resizing output predictions to a consistent dimension.
We generate 10 \algoname{} hypotheses, selecting the hypothesis with the most inliers after registration. We also report best-of-N results determined by the groundtruth to quantify the quality of the best hypothesis generated. 
Metrics are calculated after placing the estimates into the scene, first taking the mean over all objects in all images from a single scene, then the mean and standard deviation over the per-scene results. 

Results are shown in Tab. \ref{tab:bop} and Fig.~\ref{fig:bop}. Although ZeroShape exhibits good generalization performance, it can struggle when the geometry of the object is occluded, likely due to its determinism.
Using the number of inliers to select the best \algoname{} hypothesis yields better performance on average on TYO-L and NOCS.
Fig. \ref{fig:bop}a shows the relationship of taking more \algoname{} hypotheses, illustrating the utility of our multi-hypothesis method.
Using the number of inliers for hypothesis selection does not match the performance of an oracle. This is unsurprising as the registration-based metric depends only on the visible portion of the object and not the quality of the shape completion, which can result in low quality hypothesis selection.
Nevertheless, the best-of-N results show our method produces more accurate estimates on average among multiple hypotheses over all three datasets.

Finally, Fig. \ref{fig:inthewild} provides a qualitative example of shape and pose estimation on a real RGB-D image without groundtruth. 
While our results are inherently stochastic and can vary, the qualitative results show that compared to the deterministic ZeroShape which predicts overly smooth geometries and the generative Shap-E which can predict different geometries but does not have a straight-forward method of selecting the best hypothesis or pose estimation, \algoname{} is a generative method that enables pose estimation as seen in Fig. \ref{fig:inthewild}(b).

\vspace{-0.2cm}
\section{Conclusion and Limitations}
\vspace{-0.1cm}
We have presented \algoname{}, a method for shape and pose estimation in the real world. \algoname{} uses diffusion models to first predict a partial pointcloud in a Normalized Object Reference Frame, and then to predict the shape completion. 
We demonstrate that \algoname{}'s multi-hypothesis nature can lead to higher geometric accuracy considering best-of-N metrics, and the number of inliers from a registration process can be a useful signal for hypothesis selection. 

Despite \algoname{}'s promising results, limitations still exist. \algoname{} can struggle on objects with very detailed geometry and also predict incoherent surfaces (Fig. \ref{fig:ocrtoc}f). 
In practice, noisy real-world normals also appear out of distribution from synthetic training normals and can harm predictions.
Future directions include using semantic feature \cite{radford2021learning, oquab2023dinov2} conditioning, finetuning on zero-shot normal estimates \cite{bae2024dsine}, and learning a hypothesis selection metric.

\section*{ACKNOWLEDGMENT}
Code generation tools\cite{openai2024chatgpt} were used in the course of developing the code for this work.

\IEEEtriggeratref{49}
\bibliographystyle{IEEEtran}
\bibliography{references}
\end{document}